\documentclass[final]{cvpr}

\usepackage{times}
\usepackage{epsfig}
\usepackage{graphicx}
\usepackage{amsmath}
\usepackage{amsthm}
\usepackage{amssymb}
\usepackage{enumerate}
\usepackage{enumitem}
\usepackage{float}

\usepackage{multirow}

\usepackage[colorlinks]{hyperref}

\def\cvprPaperID{572} 
\def\confYear{CVPR 2021}

\begin{document}
\setlength{\abovedisplayskip}{3pt}
\setlength{\belowdisplayskip}{3pt}
\title{VIGOR: Cross-View Image Geo-localization beyond One-to-one Retrieval}

\author{Sijie Zhu, Taojiannan Yang, Chen Chen\\
Department of Electrical and Computer Engineering, University of North Carolina at Charlotte\\
{\tt\small \{szhu3, tyang30, chen.chen\}@uncc.edu}
}

\maketitle

\begin{abstract}
Cross-view image geo-localization aims to determine the locations of street-view query images by matching with GPS-tagged reference images from aerial view. Recent works have achieved surprisingly high retrieval accuracy on city-scale datasets. However, these results rely on the assumption that there exists a reference image exactly centered at the location of any query image, which is not applicable for practical scenarios. In this paper, we redefine this problem with a more realistic assumption that the query image can be arbitrary in the area of interest and the reference images are captured before the queries emerge. This assumption breaks the one-to-one retrieval setting of existing datasets as the queries and reference images are not perfectly aligned pairs, and there may be multiple reference images covering one query location. To bridge the gap between this realistic setting and existing datasets, we propose a new large-scale benchmark --VIGOR-- for cross-View Image Geo-localization beyond One-to-one Retrieval. We benchmark existing state-of-the-art methods and propose a novel end-to-end framework to localize the query in a coarse-to-fine manner. Apart from the image-level retrieval accuracy, we also evaluate the localization accuracy in terms of the actual distance (meters) using the raw GPS data. Extensive experiments are conducted under different application scenarios to validate the effectiveness of the proposed method. The results indicate that cross-view geo-localization in this realistic setting is still challenging, fostering new research in this direction. Our dataset and code will be released at \url{https://github.com/Jeff-Zilence/VIGOR}
\end{abstract}

\section{Introduction}
The objective of image-based geo-localization is to determine the location of a query image by finding the most similar image in a GPS-tagged reference database. Such technologies have proven useful for accurate localization with noisy GPS signals \cite{brosh2019accurate,zamir2010accurate} and navigation in crowded cities \cite{mirowski2018learning,li2019cross}. Recently, there has been a surge of interest in cross-view geo-localization \cite{Vo,Chen,CVM,SAFA,WACV,geocapsnet}, which uses GPS-tagged aerial-view images as reference for street-view queries. However, the performance may suffer from a large view or appearance gap between query and reference images.

Recent works \cite{CVM, SAFA, WACV} have shown that the performance of cross-view image matching can be significantly improved by feature aggregation and sample mining strategies. When the orientation of street-view (or ground-view) image is available (provided by phone-based compass), state-of-the-art methods can achieve a top-1 retrieval accuracy over 80\% \cite{SAFA}, which shows the possibility of accurate geo-localization in real-world settings. However, existing datasets \cite{Vo,Zhai,liu2019lending} simply assume that \textit{each query ground-view image has one corresponding reference aerial-view image whose center is exactly aligned at the location of the query image}. 
We argue this is not practical for real-world applications, because the query image can occur at arbitrary locations in the area of interest and the reference images should be captured before the queries emerge. In this case, perfectly aligned one-to-one correspondence is not guaranteed. 

In light of the novelty of this problem, we propose a new benchmark (VIGOR) to evaluate cross-view geo-localization in a more realistic setting. Briefly, given an area of interest (AOI), the reference aerial images are densely sampled to achieve a seamless coverage of the AOI and the street-view queries are captured at arbitrary locations. In total, $90,618$ aerial images and $238,696$ street panoramas are collected from 4 major cities in the United States (see details in Sec. \ref{sec:vigor}). The new dataset gives rise to two fundamental differences between this work and prior research.


\textbf{Beyond One-to-one:} Previous research mainly focuses on the one-to-one correspondence because existing datasets consider perfectly aligned image pairs as default. However, VIGOR enables us to explore the effect of reference samples that are not centered at the locations of queries but still cover the query area. As a result, there could be multiple reference images partially covering the same query location, breaking the one-to-one correspondence. In our geo-localization method, we design a novel hybrid loss to take advantage of multiple reference images during training.

\textbf{Beyond Retrieval:} Image retrieval can only provide image-level localization.
Since the center alignment is not guaranteed in our dataset, after the retrieval, we further employ a within-image calibration to predict the offset of the query location inside the retrieved image. Therefore, the proposed joint-retrieval-and-calibration framework provides a coarse-to-fine localization. The whole pipeline is end-to-end, and the inference is fast as the offset prediction shares the feature descriptors with the retrieval task.
Moreover, our dataset is also accompanied with raw GPS data. Thus a more direct performance assessment, \ie localization accuracy in terms of real-world distance (\eg meters), can be achieved on our dataset. 


Our main contributions can be summarized as follows:
\setlist{nolistsep}
\begin{itemize}[noitemsep,leftmargin=*]
    \item We introduce a new dataset for the problem of cross-view image geo-localization. This dataset, for the first time, allows one to study this problem under a more realistic and practical setting and offers a testbed for bridging the gap between current research and practical applications.
    \item We propose a novel joint-retrieval-and-calibration framework for accurate geo-localization in a coarse-to-fine manner, which has not been explored in the past. 
    \item We develop a new hybrid loss to learn from multiple reference images during training, which is demonstrated to be effective in various experimental settings.
    \item We also validate the potential of the proposed cross-view geo-localization framework in a real-world application scenario (assistive navigation) by simulating noisy GPS.
\end{itemize}
\section{Related Work}
\noindent\textbf{Cross-view Datasets.} A number of datasets have been proposed for cross-view geo-localization \cite{lin2013cross,workman2015wide,Vo,Zhai,Chen,liu2019lending}. Lin \etal \cite{lin2013cross} consider both satellite images and land cover attributes for cross-view geo-localization. $6,756$ ground-view images and 182,988 aerial images are collected from Charleston, South Carolina. Although the aerial images are densely sampled, they force a one-to-one  correspondence between two views and evaluation in terms of distance is not available. The original CVUSA \cite{workman2015wide} is a massive dataset containing more than $1$ million ground-level and aerial images from multiple cities in the United States. Zhai \etal \cite{Zhai} further make use of the camera's extrinsic parameters to generate aligned pairs by warping the panoramas, resulting in 35,532 image pairs for training and 8,884 image pairs for testing. This version of CVUSA is the most widely used dataset in recent research \cite{CVM,SAFA, WACV,UCF,verde2020ground} and we refer to it as CVUSA if not specified. Vo \cite{Vo} consists of about one million image pairs from 11 cities in the United States. The authors randomly collect street-view panoramas and generate several crops from each panorama along with spatially aligned aerial images from Google Maps. 
Similar to CVUSA, CVACT \cite{liu2019lending} also consists of aligned panoramas and aerial images with orientation information. It has $35,532$ image pairs for training and $92,802$ pairs for testing. In a nutshell, \emph{all these datasets consider one-to-one retrieval and none of them provide raw GPS data for localization evaluation in terms of meters}.

\noindent\textbf{Cross-view Geo-localization.}
Early works \cite{lin2013cross,workman2015wide, Vo, Chen} of cross-view geo-localization suffer from low retrieval accuracy mainly because of the significant appearance gap between two views and poor metric learning techniques. With tailored feature extractors and a modified loss function, Hu \etal \cite{CVM} show the possibility of achieving accurate localization with end-to-end deep neural networks. Several recent methods \cite{UCF,SAFA} aim to reduce the domain gap by leveraging GANs \cite{goodfellow2014generative} and polar transformations \cite{shi2020looking}. Regmi \etal \cite{UCF} propose to generate the synthetic aerial-view image from the ground-view query with a conditional GAN and adopt feature fusion to achieve better performance. SAFA \cite{SAFA} further takes advantage of the geometric prior knowledge by applying a polar transformation on the query image and replacing the global pooling with feature aggregation blocks. The top-1 accuracy of \cite{SAFA} on CVUSA \cite{Zhai} is almost 90\% if the orientation information is given.
Other approaches \cite{reweight,WACV} exploring metric learning techniques (\eg hard samples mining strategy) also show promising results on popular datasets, and they are not restricted by the geometric assumptions. \emph{However, none of these methods consider a sub-image level localization beyond the image-level retrieval or multiple reference images for training}.


\section{VIGOR Dataset}
\label{sec:vigor}
\begin{figure*}[!htbp]
    \centering
    \vspace{-0.2cm}
    \includegraphics[width=.96\linewidth]{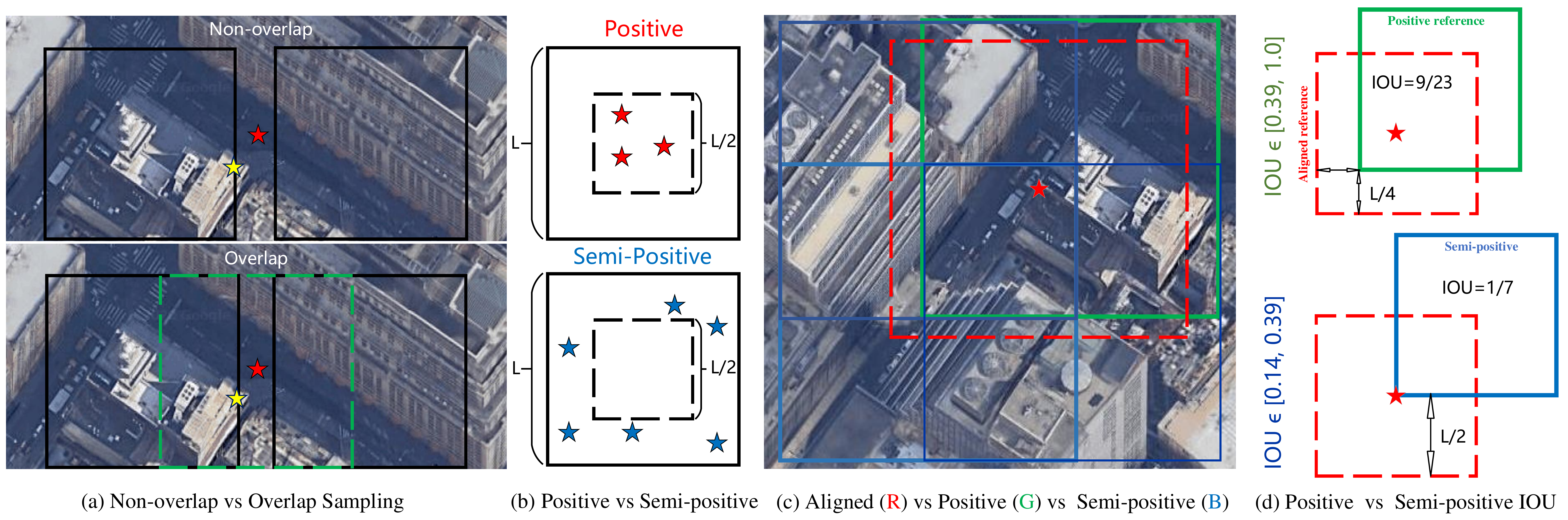}
    \vspace{-0.1cm}
    \caption{The sampling strategy of the proposed dataset. The stars denote the query locations.}
    \label{fig:overlapping_sampling}
\end{figure*}
\noindent\textbf{Problem Statement.}
Given an area of interest (AOI), our objective is to localize an arbitrary street-view query in this area by matching it with aerial reference images. To guarantee that any possible query is covered by at least one reference image, the reference aerial images must provide a seamless coverage of the AOI. As shown in Fig. \ref{fig:overlapping_sampling} (a), coarsely sampled reference images (black square boxes) are not able to provide full coverage of the AOI, and an arbitrary query location (the red star) may lie in the area between reference samples. Even if the query location (the yellow star)
lies at the edge of a reference aerial image, this reference image only shares partial (at most half) scene with the one whose center is at the query location, which may not provide enough information to be distinguished from other negative reference images. These queries can be covered by adding additional overlapping samples (the green box). 
As shown in Fig. \ref{fig:overlapping_sampling} (b), if query locations (red stars) lie at the \textbf{central area} (the black dotted box) of the $L\times L$ aerial image, the query and reference images are defined as positive samples \textit{for each other}.
Other queries (blue stars) outside the central area are defined as semi-positive samples. To guarantee that any arbitrary query has one positive reference image, we propose to densely sample the aerial images with 50\% overlap along both latitude and longitude directions as demonstrated in Fig. \ref{fig:overlapping_sampling} (c). By doing so, any arbitrary query location (the red star) in the AOI is covered by four reference images (size $L \times L$). The green box denotes the positive reference and the other three semi-positive references are denoted as blue boxes.  The positive reference is considered as ground-truth, because it has the nearest GPS to the query and contains the most shared objects with the query image. The red box denotes the perfectly aligned aerial image. Based on the definitions of positive and semi-positive as illustrated in Fig. \ref{fig:overlapping_sampling} (b), we can easily see that all positive reference images have an IOU (Intersection Over Union) greater than $0.39$ with the perfectly aligned reference (see Fig. \ref{fig:overlapping_sampling} (d)). The IOU of a typical positive sample (offset relative to the center equals to $(\pm\frac{L}{8},\pm\frac{L}{8})$) is $0.62$. The IOU between the semi-positive samples and the aligned reference falls in $[\frac{1}{7}\approx0.14, \frac{9}{23}\approx0.39]$.
\begin{figure}[!htbp]
    \centering
    \includegraphics[width=0.99\linewidth]{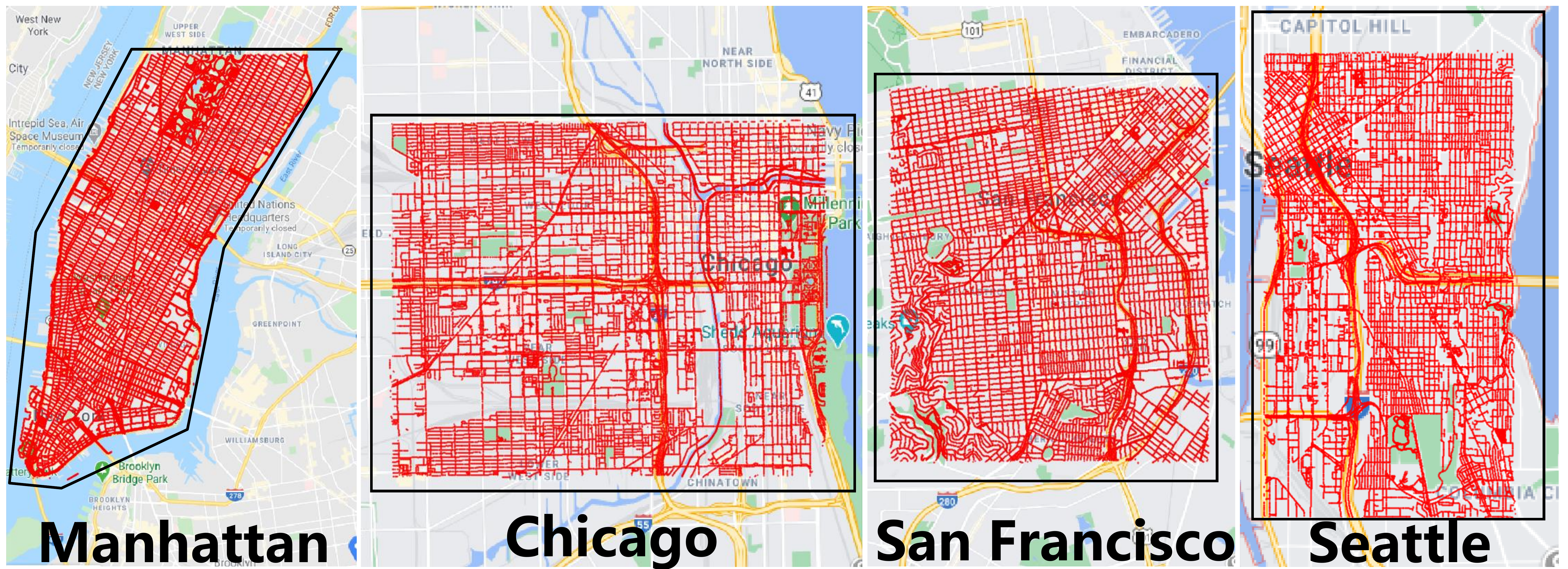}
    \caption{Aerial image coverage (black polygon) in four cities and the distributions of panoramas (red dots).}
    \label{fig:collection}
\end{figure}
\begin{figure}[!htbp]
    \centering
    \includegraphics[width=.95\linewidth]{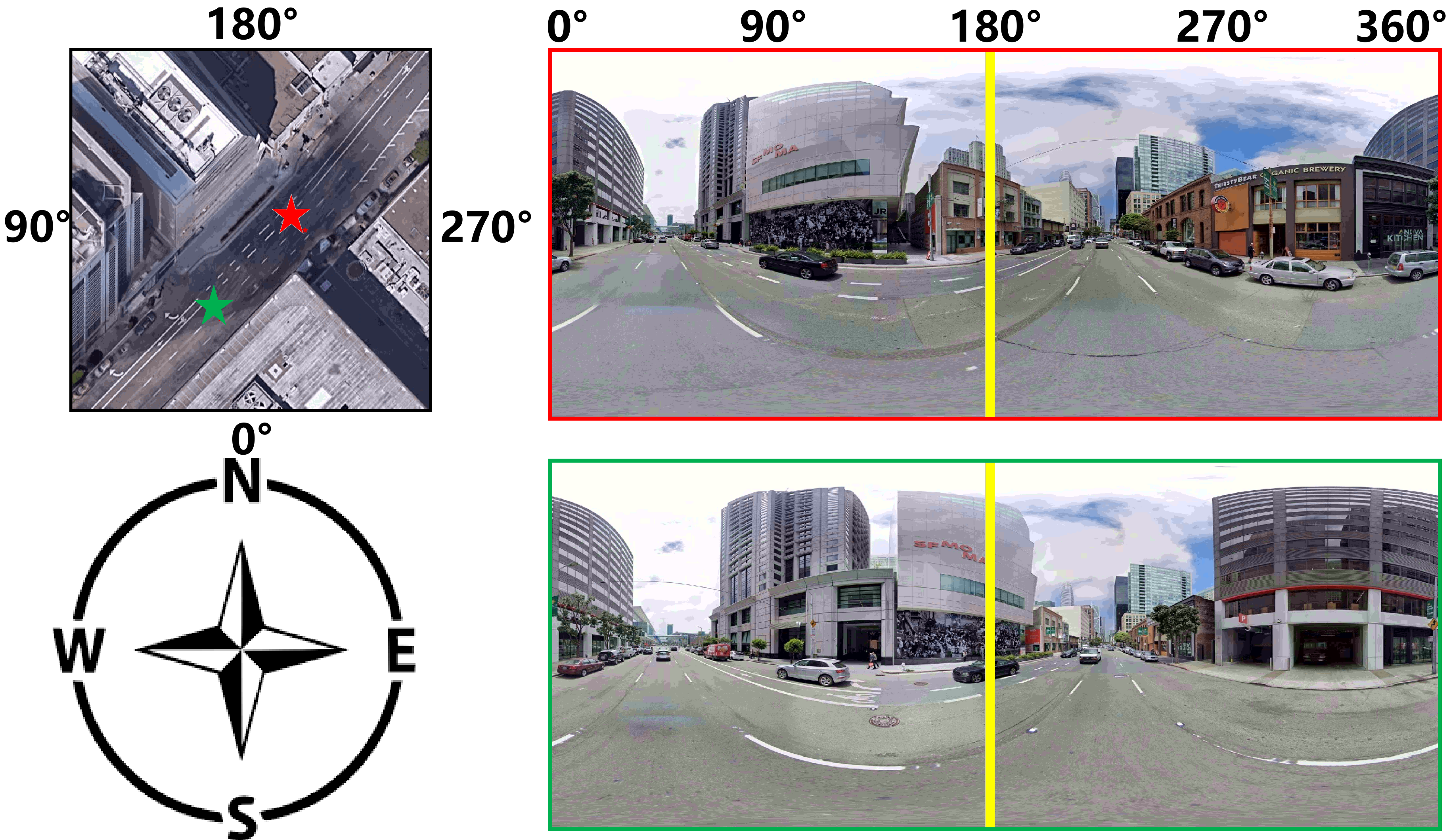}
    \caption{An example of positive samples (stars) and the orientation correspondence between aerial and ground views. The yellow bar indicates North.}
    \label{fig:positive_alignment}
\end{figure}

\noindent\textbf{Data Collection.} As shown in Fig. \ref{fig:collection}, we collect $90,618$ aerial images covering the central areas of four cities, \ie New York City (Manhattan), San Francisco, Chicago, and Seattle, as the AOI using the Google Maps Static API \cite{Google_map}. Then $238,696$ street-view panorama images are collected with the Google Street-View Static API \cite{Google_street} at zoom level 2 on most of the streets. All the GPS locations of panorama images are unique in our dataset, and the typical interval between samples is about $30$ $m$. 
We perform data balancing on the original panoramas to make sure that each aerial image has no more than 2 positive panoramas (see Fig. \ref{fig:positive_alignment}, the distributions are included in the supplementary material). This procedure results in $105,214$ panoramas for the geo-localization experiments. Also, around $4\%$ of the aerial images cover no panoramas. We keep them as distraction samples to make the dataset more realistic and challenging. The zoom level for satellite images is 20 and the ground resolution is around 0.114 $m$. The raw image sizes for aerial-view and ground-view are $640 \times 640$ and $2048 \times 1024$, respectively. Industrial-grade GPS tags for both aerial-view and ground-view images are provided for meter-level evaluation. The panoramas are then shifted according to the orientation information so that North lies in the middle. Fig. \ref{fig:positive_alignment} shows an example of orientation correspondence between a pair of aerial and street-view images.
\begin{table*}[htbp]
\small
    \centering
    \begin{tabular}{l|c|c|c|c}
    \hline
    
    \hline
    & Vo \cite{Vo} & CVACT \cite{liu2019lending} &  CVUSA \cite{Zhai} & VIGOR (proposed)\\
    \hline 
    Satellite images& $\sim 450,000$ & $128,334$ & $44,416$  & $90,618$ \\
    Panoramas in total & $\sim 450,000$ & $128,334$ & $44,416$ & $238,696$ \\
    Panoramas after balancing & - & - & - & $105,214$\\
    Street-view GPS locations & Aligned & Aligned & Aligned & Arbitrary \\
    Full panorama & No & Yes & Yes & Yes\\
    Multiple cities & Yes & No &  Yes  & Yes\\
    Orientation information & Yes & Yes & Yes & Yes\\
    Evaluation in terms of meters & No & No & No & Yes \\
    Seamless coverage on area of interest & No & No &No & Yes\\
    Number of references covering each query & 1 & 1 & 1 & 4 \\
    \hline
    
    \hline
    \end{tabular}
    \caption{Comparison between the proposed VIGOR dataset and existing datasets for cross-view geo-localization.} 
    \label{tab:dataset}
\end{table*}

\noindent\textbf{Head-to-head Comparison.} Table~\ref{tab:dataset} shows a comparison between our dataset and previous benchmarks. The most widely used dataset, CVUSA \cite{Zhai}, consists of images mainly collected at suburban areas. Our dataset, on the other hand, is collected for urban environments. In practice, the GPS signal is more likely to be noisy in urban areas than suburban (\eg the phone-based GPS error can be up to 50 meters in Manhattan \cite{brosh2019accurate}). 
Therefore, our dataset has more potential application scenarios, \eg vision-based mobile assistive navigation. Besides, urban areas are more crowded with tall buildings. The mutual semantics between ground and aerial views are significantly reduced by occlusions and shadows, making our dataset more challenging than CVUSA. 
Furthermore, previous datasets simply adopt one-to-one retrieval for evaluation, which is not the case of real-world scenarios, because it is impossible to predict the location of an arbitrary query and capture an aligned reference image there beforehand. Our dataset considers arbitrary query locations, and even the ground-truth reference image does not have the same GPS location as the query; thus it is more realistic but challenging for retrieval. Our dataset also provides the raw GPS data for meter-level evaluation which is the ultimate goal of localization applications. We believe that our dataset is a great complement to the existing cross-view image datasets, and can be served as a testbed for bridging the gap between current research and practical applications.
\begin{table}[h!]
\small
    \centering
    
    \begin{tabular}{l|c|c|c|c|c}
    \hline
    
    \hline
     \multicolumn{2}{c|}{} & \multicolumn{2}{c}{Same-Area}& \multicolumn{2}{|c}{Cross-Area} \\
    \cline{3-6}
     \multicolumn{2}{c|}{} & Number & City & Number & City\\
     \hline
     \multirow{2}{1.7em}{Train} & Aerial  & 90,618 & All & 44,055 & \multirow{2}{5.7em}{New York Seattle}\\
     \cline{2-5}
     ~ & Street & 52,609 & All &  51,520 &  \\
     \cline{1-6}
     \multirow{2}{1.7em}{Test} & Aerial & 90,618 & All & 46,563 & \multirow{2}{5.7em}{San Francisco Chicago}\\
     \cline{2-5}
     ~ & Street & 52,605 & All & 53,694 & \\
    \hline
    
    \hline
    \end{tabular}
    \caption{The evaluation splits of VIGOR in two settings.}
    \label{tab:split}
\end{table}

\noindent\textbf{The Evaluation Protocol.} 
We design two evaluation settings for the experiments, \ie same-area and cross-area evaluation, according to different application scenarios.\\
\textbf{Same-area}: If one plans to build an aerial-view reference database for arbitrary street queries in an AOI, the goal of model training is to handle arbitrary new queries. Therefore, the best solution would be collecting GPS-tagged queries in the same area for training rather than training in other areas with cross-area transfer. In this case, the aerial images in four cities are all included as the reference data for both training and testing. Then all the street panoramas are randomly split into two disjoint sets (see Table \ref{tab:split}).\\
\textbf{Cross-area}: For cities where no GPS-tagged queries are available for training, the cross-area transfer is necessary. For this setting, all the images from New York and Seattle are used for training, and images from San Francisco and Chicago are held out for evaluation.

\section{Coarse-to-fine Cross-view Localization}
In this section, we propose a joint-retrieval-and-calibration framework for geo-localization in a coarse-to-fine manner. Sec. \ref{sec:baseline} introduces a strong baseline built with state-of-the-art techniques based on \textit{only the positive samples}. Sec. \ref{sec:hybrid} proposes an IOU-based semi-positive assignment loss to leverage the supervision information of semi-positive samples. With the retrieved best matching reference image, Sec. \ref{sec:offset} aims to estimate the offset of the query GPS location relative to the center of the retrieved aerial-view image as a meter-level calibration.

\subsection{Baseline Framework}
\label{sec:baseline}
To achieve satisfactory results on the proposed dataset, it is important to adopt state-of-the-art techniques to build a strong baseline. Therefore, we employ the feature aggregation module of SAFA (spatial-aware feature aggregation) \cite{SAFA} with the global negative mining strategy from \cite{WACV}.
\begin{figure*}
    \centering
    \vspace{-0.2cm}
    \includegraphics[width=.9\linewidth]{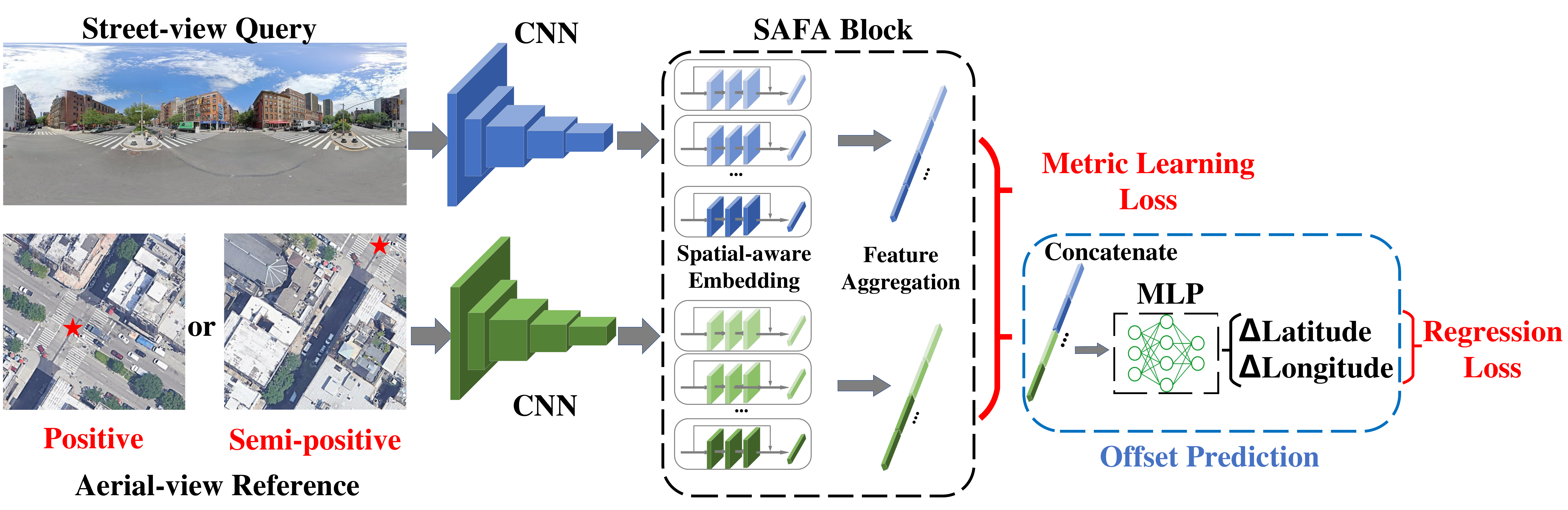}
    \vspace{-0.1cm}
    \caption{An overview of the proposed end-to-end framework. The Siamese network provides embedding features for retrieval as a coarse image-level localization. The offset prediction further generates refined localization in terms of meters.}
    \label{fig:arc}
\end{figure*}

\noindent\textbf{Feature Aggregation.} SAFA \cite{SAFA} is a combination of polar transformation, Siamese backbone and feature aggregation blocks. However, the polar transformation assumes that the ground-view GPS is at the center of the corresponding aerial-view reference image, which does not apply in our case. Therefore, we only adopt the feature aggregation in our framework (see Fig. \ref{fig:arc}). The main idea of the feature aggregation block is to re-weight the embeddings in accordance with their positions. The spatial-aware block provides a significant performance gain when the orientation information of query images is available.\\ 
\noindent\textbf{Mining Strategy.} Metric learning literature \cite{facenet,suh2019stochastic,metricreality} has revealed the importance of mining hard samples during training, as the model would suffer from poor convergence when most samples barely contribute to the total loss. For cross-view geo-localization, \cite{WACV} further shows the importance of mining global hard samples instead of mining within a mini-batch. The key idea is to build a first-in-first-out mining pool to cache the embedding of the hardest sample and refresh the pool along with back propagation efficiently. In a mini-batch, the first half images are randomly selected and the global hard samples with respect to each of them are mined from the mining pool to form the other half of the batch. We adopt this efficient global mining strategy \cite{WACV} in the baseline to further improve its performance.

\subsection{IOU-based Semi-positive Assignment}
\label{sec:hybrid}
If we only consider the positive samples, the retrieval problem can be tackled with standard metric learning. For the baseline, we adopt the widely used triplet-like loss proposed in \cite{CVM}:
\begin{equation}
    \mathcal{L}_{triplet} = log\left(1+e^{\alpha(d_{pos}-d_{neg})}\right),
    \label{eq:triplet}
\end{equation}
where $d_{pos}$ and $d_{neg}$ denote the squared $l_{2}$ distance of the positive and negative pairs. In a mini-batch with $N$ ground-view and aerial-view image pairs, we use the exhaustive strategy \cite{facenet} to build $2N(N-1)$ triplets, thereby making full use of all the input images. Following \cite{CVM}, we adopt $l_{2}$ normalization on the output embedding features.

In addition to positive samples, it can be beneficial to take advantage of the supervision information of semi-positive samples. However, simply assigning semi-positive samples as positive would hurt the retrieval performance. For a street-view query, the semi-positive aerial images only contain a small part of the scene at the query location, thus the similarities in the feature embedding space between semi-positive samples and the query should not be as high as those of positive samples. An intuitive idea is to assign the similarity according to the IOU between the reference image and the aligned one (see Fig. \ref{fig:overlapping_sampling} (d)). Therefore, the IOU-based semi-positive assignment loss is expressed as:
\begin{equation}
    \mathcal{L}_{IOU}=\left(\frac{S_{semi}}{S_{pos}}-\frac{IOU_{semi}}{IOU_{pos}}\right)^{2},
    \label{eq:iou}
\end{equation}
where $S_{pos}$ and $S_{semi}$ denote the cosine similarity of the positive and semi-positive pairs in the embedding space. $IOU_{pos}$ and $IOU_{semi}$ denote the IOU of positive and semi-positive pairs. This loss forces the ratio of the similarities in the embedding space to be close to the ratio of IOUs. Other assignment strategies for the semi-positive samples are also investigated and evaluated in the ablation study.

\subsection{Offset Prediction}
\label{sec:offset}
With the top-1 retrieved reference aerial image, we employ an auxiliary task to further refine the localization inside the aerial-view image in a unified framework (see Fig. \ref{fig:arc}). With image retrieval, the minimal interval between retrieved reference images in our dataset is half of the width of aerial images ($L/2$). To achieve more fine-grained localization, we apply an MLP (Multilayer Perceptron) to predict the offset of the query location relative to the center of the retrieved reference image. As shown in Fig. \ref{fig:arc}, the auxiliary MLP consists of two fully connected layers and takes the concatenated embedding features as input. Here we use regression to generate the prediction, while we also provide a comparison with classification in the experiments. The offset regression loss is formulated as:
\begin{equation}
    \mathcal{L}_{offset} = (lat-lat^{*})^{2} + (lon-lon^{*})^{2},
    \label{eq:offset}
\end{equation}
where $lat$ and $lon$ denote the predicted offset of the query GPS location relative to the reference GPS in latitude and longitude directions, and $lat^{*}$ and $lon^{*}$ denote the ground-truth offset. They are all converted into meters and normalized with $L$ during training. The final hybrid loss function is given by:
\begin{equation}
    \mathcal{L}_{hybrid}=\mathcal{L}_{triplet}+\mathcal{L}_{IOU}+\mathcal{L}_{offset}. 
    \label{eq:hybrid}
\end{equation}


\section{Experiments}
\begin{table*}[!htbp]
\small
    \centering
    \begin{tabular}{r|c|c|c|c|c|c|c|c}
    \hline
    
    \hline
    \multirow{2}{4em}{} & \multicolumn{4}{c}{Same-Area}& \multicolumn{4}{|c}{Cross-Area} \\
    \cline{2-9}
      ~ & Top-1 & Top-5 & Top-1\% & Hit Rate & Top-1 & Top-5 & Top-1\% & Hit Rate\\
     \hline
     Siamese-VGG ($\mathcal{L}_{triplet}$) & 18.1 & 42.5 & 97.5 & 21.2 & 2.7 & 8.2 & 61.7 & 3.1\\
     SAFA ($\mathcal{L}_{triplet}$) & 33.9 & 58.4 & 98.2 & 36.9 & 8.2 & 19.6 & 77.6 & 8.9\\
     SAFA+Mining (baseline, $\mathcal{L}_{triplet}$) & 38.0 & 62.9 & 97.6 & 41.8 & 9.2 & 21.1 & 77.8 & 9.9\\
     Ours ($\mathcal{L}_{hybrid}$) & \textbf{41.1} & \textbf{65.8} & \textbf{98.4} & \textbf{44.7} & \textbf{11.0} & \textbf{23.6} & \textbf{80.2} & \textbf{11.6}\\
    \hline
    
    \hline
    \end{tabular}
    \caption{Retrieval accuracy (percentage) of different methods.}
    \label{tab:benchmarking}
\end{table*}
\begin{figure*}[!htbp]
\centering
\vspace{-0.2cm}
\includegraphics[width=0.45\linewidth]{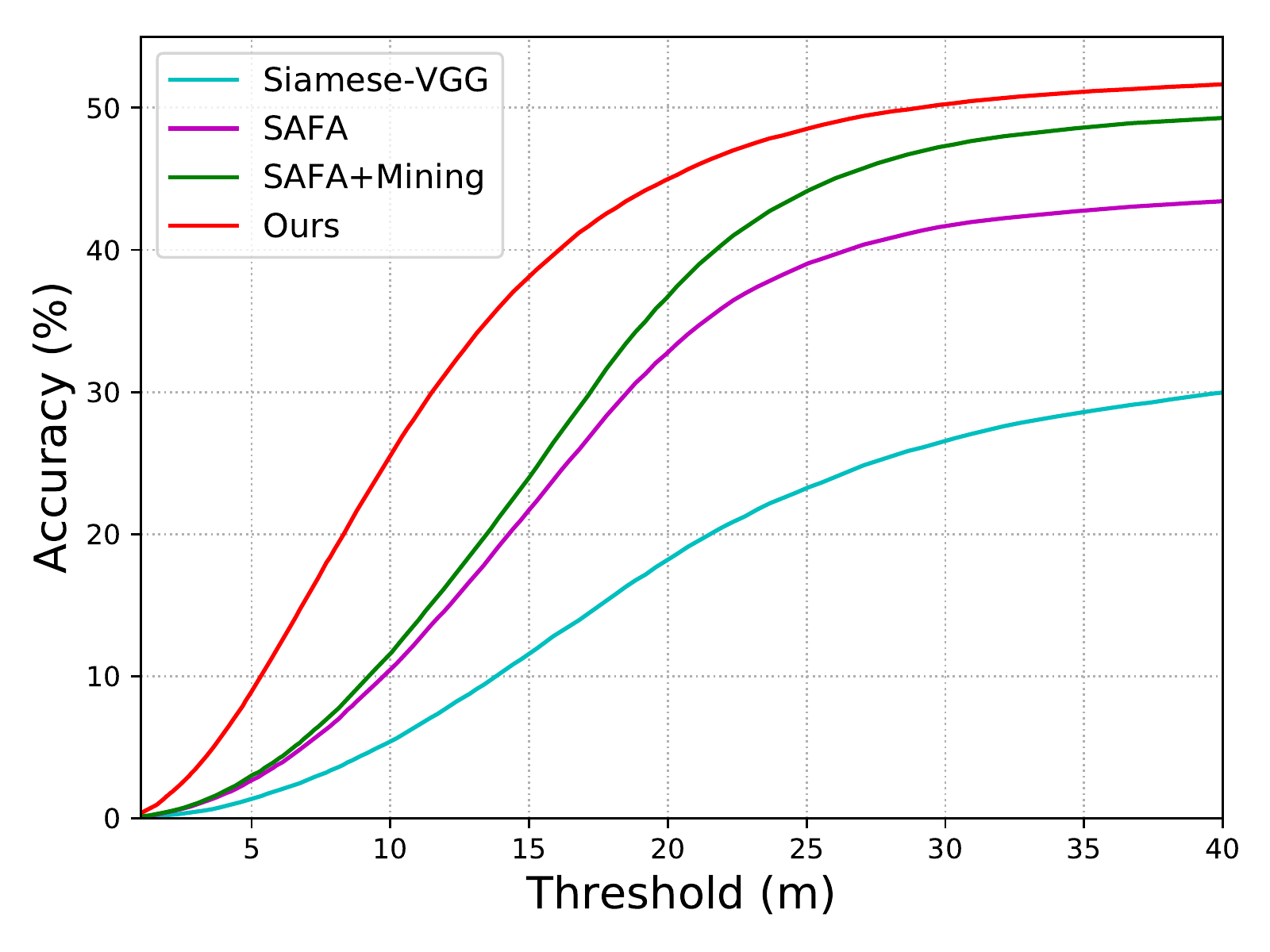}
\includegraphics[width=0.45\linewidth]{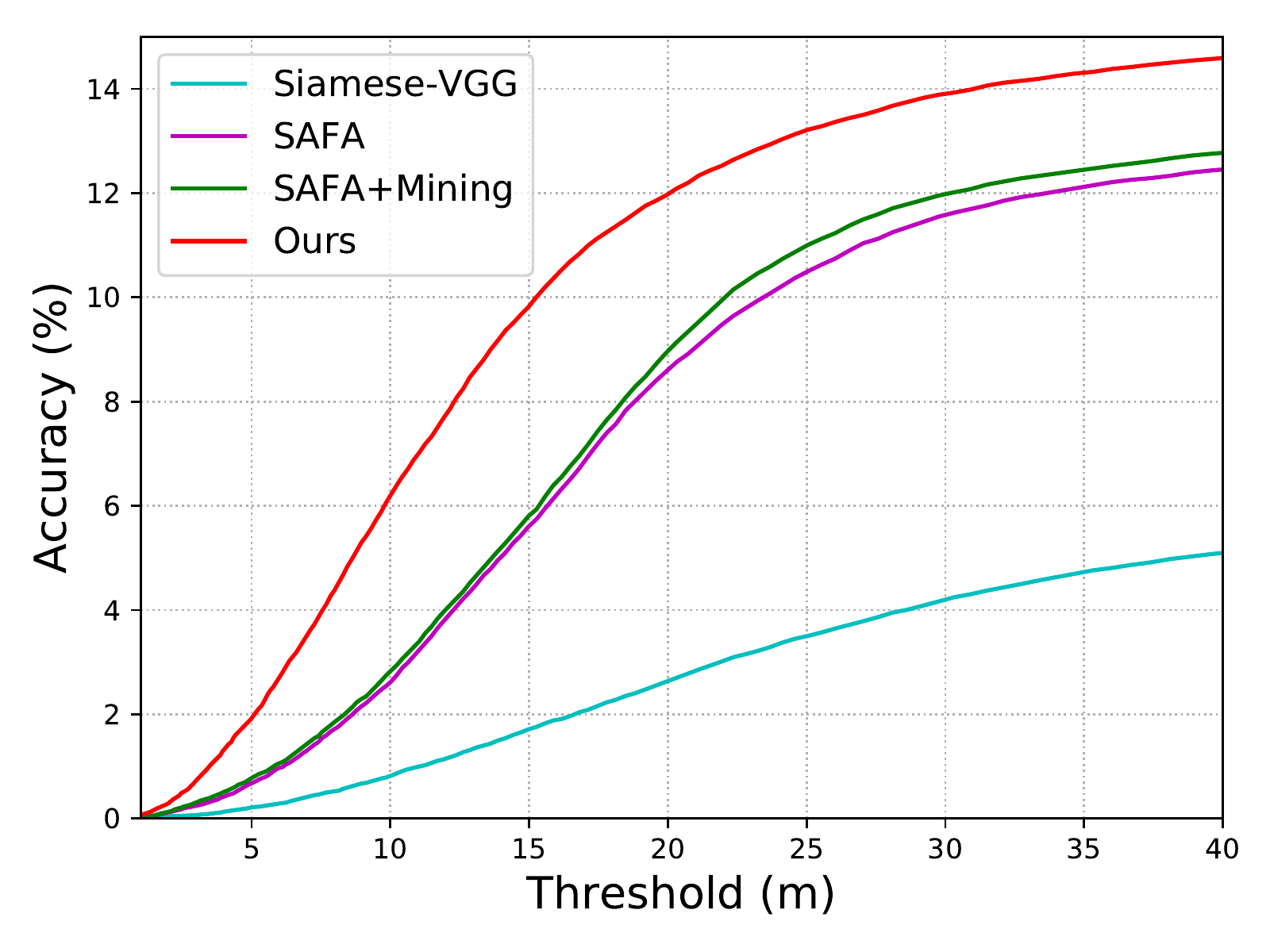}
\vspace{-0.3cm}
\caption{Same-area (left) and cross-area (right) meter-level localization accuracy of different methods.}
\label{fig:benchmark}
\end{figure*}
\begin{table*}[!htbp]
\small
    \centering
    \begin{tabular}{r|c|c|c|c|c|c|c|c}
    \hline
    
    \hline
    \multirow{2}{12em}{Semi-positive Assignment} & \multicolumn{4}{c}{Same-Area}& \multicolumn{4}{|c}{Cross-Area} \\
    \cline{2-9}
      ~ & Top-1 & Top-5 & Top-1\% & Hit Rate & Top-1 & Top-5 & Top-1\% & Hit Rate\\
     \hline
     No semi-positive (\ie baseline, $\mathcal{L}_{triplet}$) & 38.0 & 62.9 & 97.6 & 41.8 & 9.2 & 21.1 & 77.8 & 9.9\\
     Positive ($\mathcal{L}_{triplet}$) & 20.3 & 45.7 & 97.9 & 25.4 & 2.7 & 7.6 & 58.2 & 3.1\\
     IOU ($\mathcal{L}_{triplet}$+$\mathcal{L}_{IOU}$) & \textbf{41.1} & \textbf{65.9} & 98.3 & \textbf{44.8} & \textbf{10.7} & \textbf{23.5} & \textbf{79.3} & \textbf{11.4}\\
     Positive+IOU ($\mathcal{L}_{triplet}$+$\mathcal{L}_{IOU}$) & 31.1 & 58.3 & \textbf{98.6} & 36.7 & 5.3 & 13.6 & 69.4 & 6.0 \\
    \hline
    
    \hline
    \end{tabular}
    \caption{Retrieval accuracy (percentage) of the proposed method with different semi-positive assignment strategies.}
    \label{tab:hybrid}
\end{table*}
\noindent\textbf{Implementation Details.} All the experiments are deployed based on Tensorflow \cite{abadi2016tensorflow}. Ground-view panoramas and aerial-view images are resized to $640\times 320$ and $320\times 320$ respectively before being fed into the network. VGG-16 \cite{vgg} is adopted as the backbone feature extractor and 8 SAFA blocks are used by following \cite{SAFA}. The mining strategy parameters are set the same as in \cite{WACV}. 
Following \cite{CVM}, we set $\alpha$ in the $\mathcal{L}_{triplet}$ loss to 10. The Adam optimizer \cite{adam} is used with a learning rate of $10^{-5}$. 
Our method is first trained with $\mathcal{L}_{triplet}$. Then it switches to the hybrid loss (Eq. \ref{eq:hybrid}) after 30 epochs for the same-area setting and 10 epochs for the cross-area setting.
The baseline (Sec.~\ref{sec:baseline}) for comparison is only trained with $\mathcal{L}_{triplet}$. 
\\ \noindent\textbf{Evaluation Metrics.} We first evaluate the retrieval performance with the top-$k$ recall accuracy following previous works \cite{CVM,SAFA}. For each test query, its closest $k$ reference neighbors in the embedding space are retrieved as prediction. One retrieval is considered correct if the ground-truth image is included in the top-$k$ retrieved images. If the retrieved top-1 reference image covers the query image  (including the ground-truth), it is considered as a hit and the hit rate is also provided for retrieval evaluation. Moreover, we compute the real-world distance between the top-1 predicted location and the ground-truth query GPS as meter-level evaluation. 
\\ \noindent\textbf{Main Results.}
On the proposed VIGOR dataset, we compare the proposed method with previous approaches under both same-area and cross-area settings. ``Siamese-VGG'' \cite{CVM} is a simple Siamese-VGG network with global average pooling, and is trained with $\mathcal{L}_{triplet}$. ``SAFA'' and ``SAFA + mining'' denote the SAFA \cite{SAFA} architecture w/o and w/ the mining strategy in \cite{WACV} using $\mathcal{L}_{triplet}$. As shown in Table \ref{tab:benchmarking} and Fig. \ref{fig:benchmark}, the proposed method constantly outperforms previous approaches in terms of both retrieval and meter-level evaluation. Compared with ``SAFA+Mining" (the baseline), the relative improvements of our method for the 10-meter-level accuracy (see Fig. \ref{fig:benchmark}) are $124\%$ ($11.4\%$ $\rightarrow$ $25.5\%$) in the same-area setting, and $121\%$ ($2.8\%$	$\rightarrow$ $6.2\%$) in the cross-area setting. The substantial improvements reveal the superiority of the proposed hybrid loss.
\begin{table*}[htbp]
\small
    \centering
    \begin{tabular}{c|c|c|c|c|c|c|c|c}
    \hline
    
    \hline
    \multirow{2}{8em}{Offset Prediction} & \multicolumn{4}{c}{Same-Area}& \multicolumn{4}{|c}{Cross-Area} \\
    \cline{2-9}
      ~ & Top-1 & Top-5 & Top-1\% & Hit Rate & Top-1 & Top-5 & Top-1\% & Hit Rate\\
     \hline
     None (retrieval-only) & 41.1 & 65.9 & 98.3 & 44.8 & 10.7 & 23.5 & 79.3 & 11.4\\
     Regression & 41.1 & 65.8 & \textbf{98.4} & 44.7 & \textbf{11.0} & \textbf{23.6} & \textbf{80.2} & \textbf{11.6}\\
     Classification & \textbf{41.5} & \textbf{66.3} & \textbf{98.4} & \textbf{45.2} & 10.7 & 23.2 & 79.3 & 11.4\\
    \hline
    
    \hline
    \end{tabular}
    \caption{Retrieval accuracy (percentage) of the proposed method with different offset prediction schemes.}
    \label{tab:offset}
\end{table*}
\begin{figure*}[!]
\centering
\vspace{-0.25cm}
\includegraphics[width=0.44\linewidth]{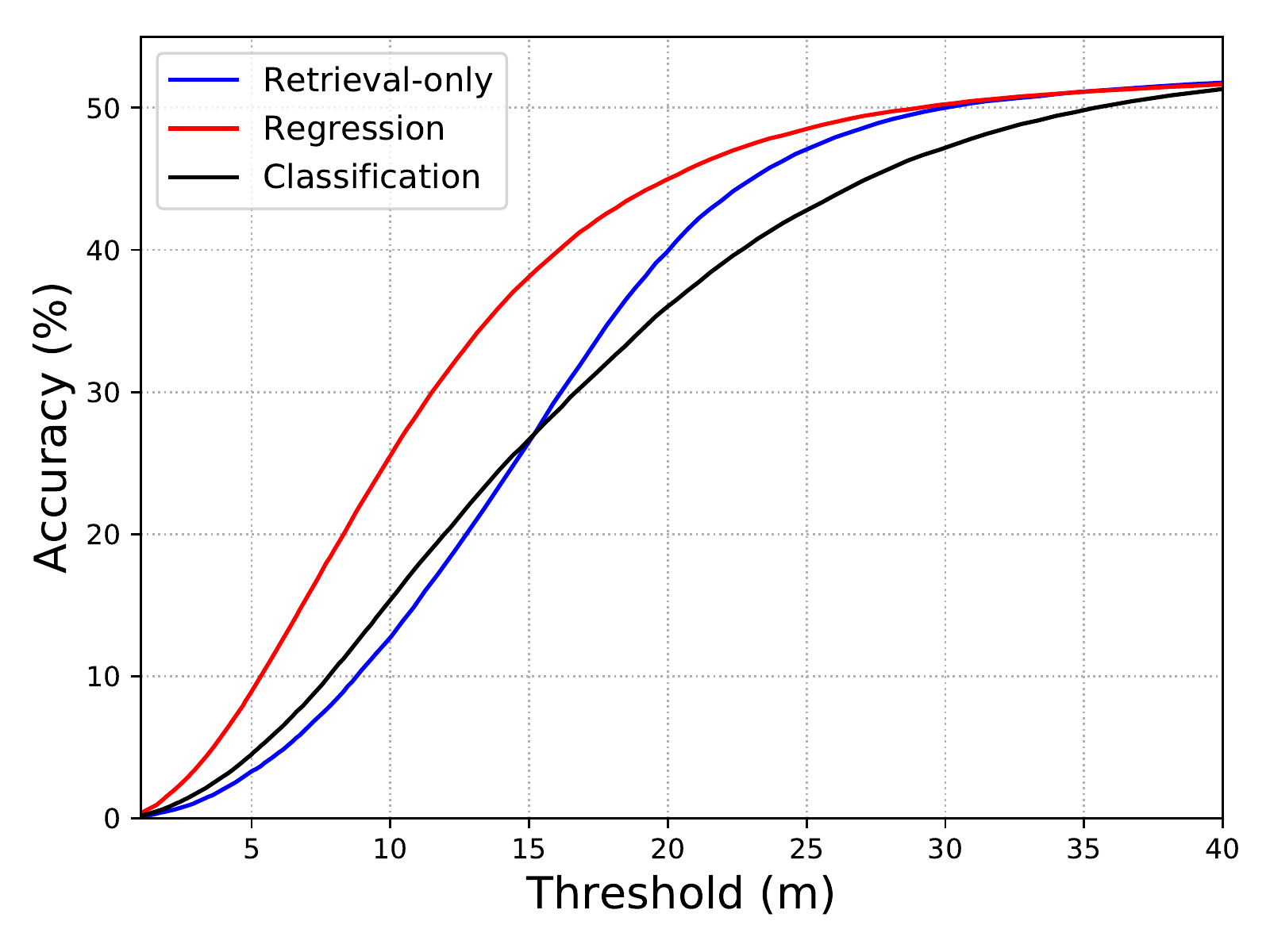}
\includegraphics[width=0.44\linewidth]{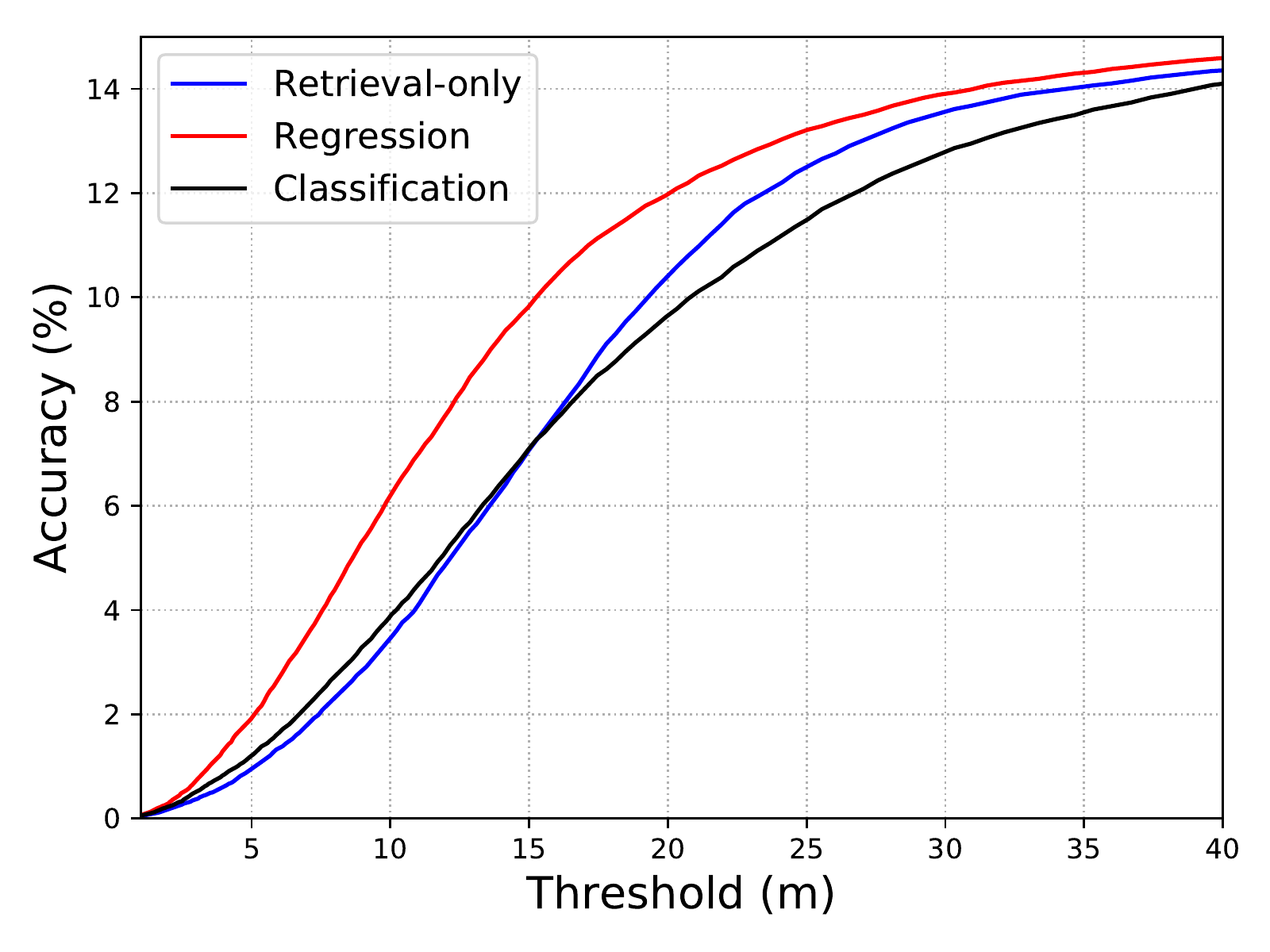}
\vspace{-0.3cm}
\caption{Same-area (left) and cross-area (right) meter-level localization accuracy of different offset prediction methods.}
\label{fig:offset}
\end{figure*}
\section{Ablation Study}
\noindent\textbf{Semi-positive Assignment.}
We compare four semi-positive assignment strategies.
``No semi-positive" denotes the baseline which ignores the semi-positive samples.
``Positive'' means assigning semi-positive samples as positive and using $\mathcal{L}_{triplet}$ (Eq.~\ref{eq:triplet}). ``IOU'' denotes our IOU-based assignment (Eq.~\ref{eq:iou}). ``Positive+IOU'' means including the semi-positive samples as positive in $\mathcal{L}_{triplet}$ along with the IOU-based assignment loss. The results in Table \ref{tab:hybrid} show that only IOU-based assignment (``IOU") boosts the performance compared with the baseline. Based on the results of ``Positive" and ``Positive+IOU", assigning semi-positive samples as positive hinders the retrieval performance whether the IOU-based loss is used or not.

\begin{figure}[htbp]
    \centering
    \vspace{-0.1cm}
    \includegraphics[width=.95\linewidth]{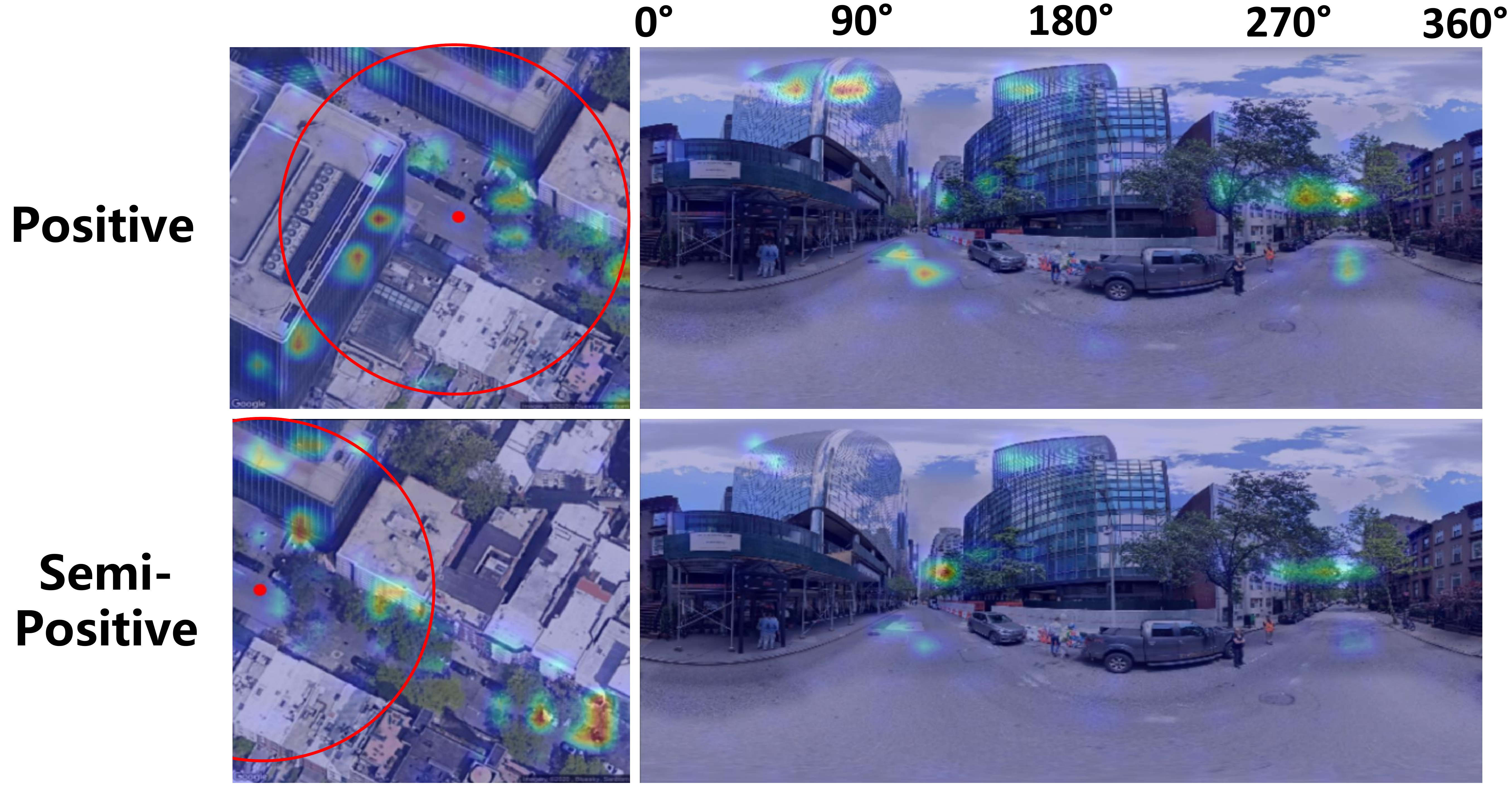}
    \caption{Visualization results of the query image matched with positive and semi-positive reference aerial images. Red circle denotes the query region.}
    \label{fig:rgradcam}
\end{figure}

\vspace{-0.1cm}
To further illustrate the difference between positive and semi-positive matching, we conduct \textbf{visual explanation}. Specifically, we use Grad-CAM \cite{selvaraju2017grad,zhu2019visual} to show which regions contribute the most to the cosine similarity of the embedding features of two views. As presented in Fig. \ref{fig:rgradcam}, for a query image, we select the ground-truth reference aerial image (\ie positive) and a semi-positive image (the query GPS location lies at its edge area), and generate the activation maps of both views. For the positive matching case, the surrounding objects (buildings, roads and trees) are all available to provide high contribution to the similarity between two views. However, in the case of semi-positive matching, two views only share half of the scene and the building on the west of the query (around $90^{\circ}$ in panorama) does not contribute to the similarity, because it is not in this semi-positive image. This example shows how the intersection semantics of two views affect the image matching, which agrees with the design of our IOU-based assignment. 
\begin{figure}[htbp]
    \centering
    \vspace{-0.2cm}
    \includegraphics[width=.95\linewidth]{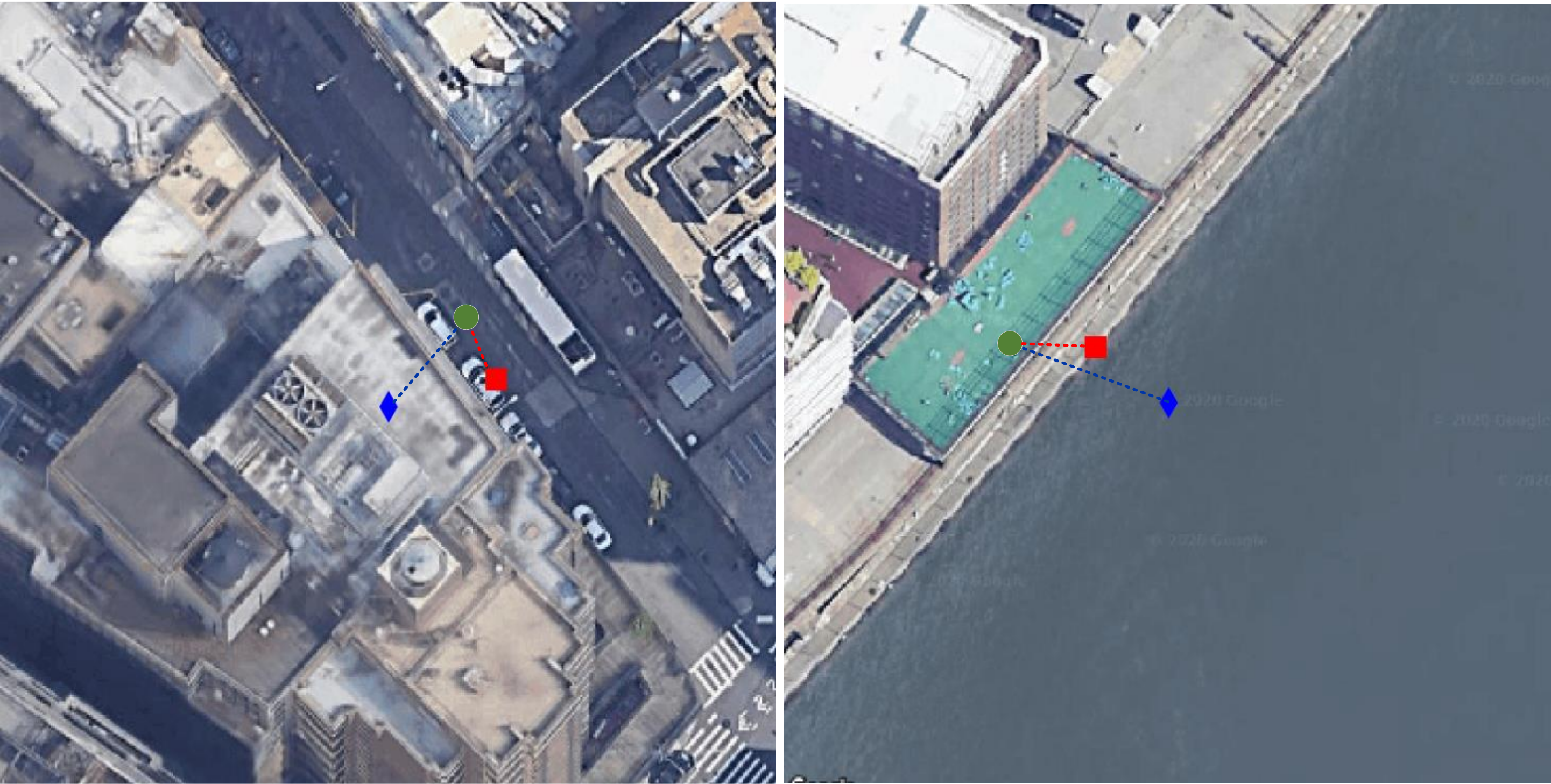}
    \caption{Case study on meter-level refinement within the retrieved aerial image. Red square, green circle and blue diamond denote the final prediction with regression, ground-truth, and center (\ie the prediction with only retrieval), respectively.} 
    \vspace{-0.5cm}
    \label{fig:cases}
\end{figure}
\vspace{-0.1cm}
\\ \noindent\textbf{Offset Prediction.} 
We investigate both regression and classification for offset prediction in our method. For classification, we split the central area of an aerial image (offset in $[-L/4,L/4]$) into a $10\times 10$ grid, leading to 100 classes for classification. As shown in Table \ref{tab:offset}, both regression and classification have negligible improvement on the retrieval performance. However, as evident from Fig. \ref{fig:offset}, regression-based calibration significantly boosts the meter-level accuracy in both settings. For example, the regression method almost doubles the 10-meter-level localization accuracy. However, classification does not work well for calibration possibly due to the ambiguous supervision of grid-based classification. We provide a case study in Fig. \ref{fig:cases} to show examples of
predicted offset on aerial images. 
\begin{table*}[!htbp]
\small
    \centering
    \begin{tabular}{c|c|c|c|c|c|c|c|c|c}
    \hline
    
    \hline
    \multirow{2}{4em}{}& \multirow{2}{6em}{Positive per Aerial Image} & \multicolumn{4}{c}{Same-Area}& \multicolumn{4}{|c}{Cross-Area} \\
    \cline{3-10}
      ~ & ~& Top-1 & Top-5 & Top-1\% & Hit Rate & Top-1 & Top-5 & Top-1\% & Hit Rate\\
     \hline
     \multirow{2}{4em}{Baseline} & 2  & 38.0 & 62.9 & 97.6 & 41.8 & 9.2 & 21.1 & 77.8 & 9.9\\
     ~& 3 & 46.0 & 70.8 & 98.5 & 50.8 & 10.6 & 23.5 & 79.5 & 11.5\\
     \hline
     \multirow{2}{4em}{Ours} & 2  & 41.1 & 65.9 & 98.3 & 44.8 & 10.7 & 23.5 & 79.3 & 11.4\\
     ~& 3 & 48.5 & 72.9 & 98.9 & 52.6 & 11.5 & 24.8 & 80.8 & 12.2\\
    \hline
    
    \hline
    \end{tabular}
    \caption{Retrieval accuracy (percentage) of the proposed method with different sample balancing settings.}
    \label{tab:balancing}
\end{table*}
\begin{figure*}[!htbp]
\centering
\vspace{-0.2cm}
\includegraphics[width=0.4\linewidth]{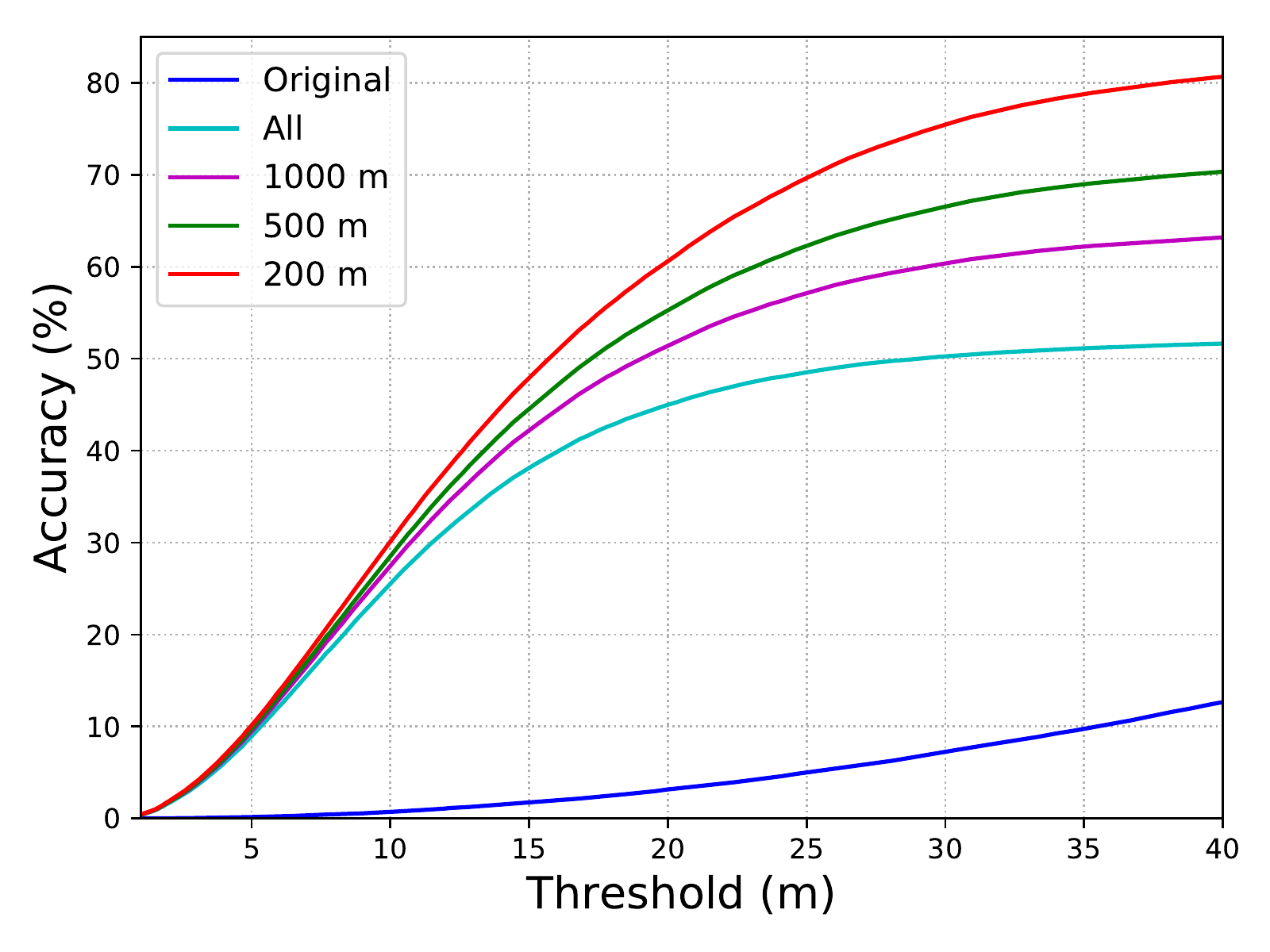}
\includegraphics[width=0.4\linewidth]{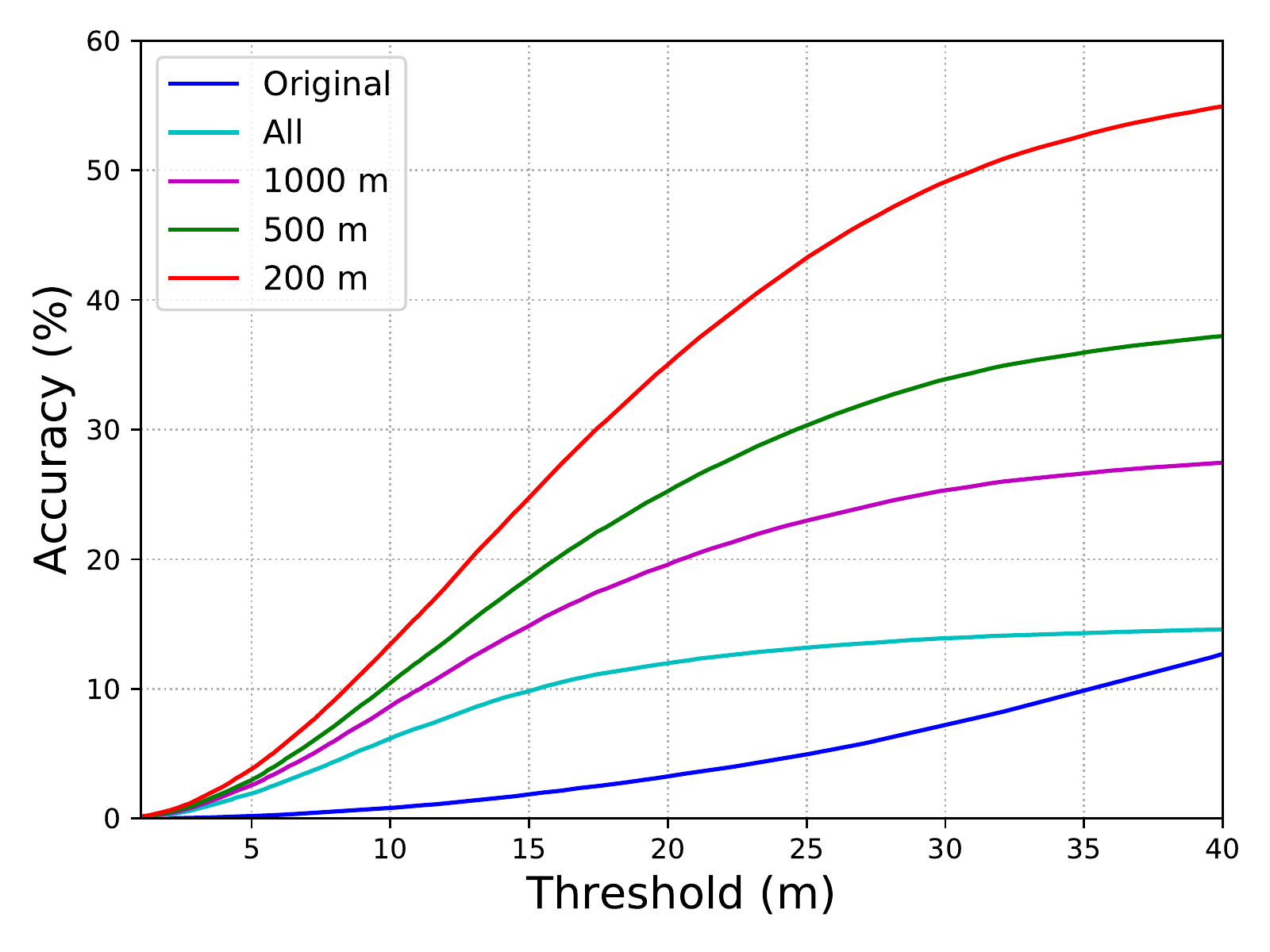}
\vspace{-0.2cm}
\caption{Same-area (left) and cross-area (right) meter-level localization accuracy of different search scopes given noisy GPS signal.}
\label{fig:noisy}
\end{figure*}

\noindent\textbf{Sample Balancing.} To investigate the effect of sample balancing in the pre-processing procedure, we compare ``balancing-2" with ``balancing-3" in Table \ref{tab:balancing_infor}. The results in Table \ref{tab:balancing} show that more densely sampled panoramas bring slightly better performance on both settings, while the improvement is consistent across different balancing settings.
\vspace{-0.4cm}
\begin{table}[!htbp]
    \centering
    \small
    \begin{tabular}{r|c|c}
    \hline
    
    \hline
     ~& Balancing-2 & Balancing-3\\
     \hline
     Positives per aerial image & 2 & 3\\
     Number of panoramas & 105,214 & 149,869 \\
     Number of aerial images & 90,618 & 90,618\\
    \hline
    
    \hline
    \end{tabular}
    \caption{The proposed dataset with different balancing numbers.}
    \label{tab:balancing_infor}
\end{table}
\vspace{-0.4cm}
\section{Application: Assistive Navigation}
\label{sec:noisygps}
The GPS data provided by commercial devices such as phones could be noisy in urban environments (\eg the phone-based GPS error can reach up to 50 meters in Manhattan \cite{brosh2019accurate}). Image geo-localization can assist mobile navigation. 
To further validate the potential of cross-view geo-localization given noisy GPS signals \cite{brosh2019accurate,zamir2010accurate}, we simulate noisy GPS signals by adding random offsets in $[-100 m,100 m]$ to the ground-truth GPS data (latitude and longitude) in our dataset. \textbf{In the inference stage}, the query image can be matched with only a small sub-set of reference images around the noisy GPS locations instead of the entire reference database (denoted by ``All"). For a noise level of $100 m$, a search scope of $200m$ 
is sufficient to cover all possible references. To better illustrate the navigational assistance provided by our image geo-localization, we compare the results of multiple scopes with simply using the noisy GPS signals (``Original"). As shown in Table \ref{tab:noisy} and Fig. \ref{fig:noisy}, smaller search scopes generate better results for both retrieval and meter-level evaluation because there are less negative reference samples. The same-area evaluation even yields an accuracy higher than $70\%$ for $30m$-level localization. Moreover, as compared to the original noisy GPS, our cross-view geo-localization method significantly improves the localization accuracy, demonstrating its practicality in real-world applications. 
\vspace{-0.2cm}
\begin{table}[!htbp]
    \centering
    \small
    \begin{tabular}{c|c|c|c|c}
    \hline
    
    \hline
    \multirow{2}{6em}{Search Scope} & \multicolumn{2}{c}{Same-Area}& \multicolumn{2}{|c}{Cross-Area} \\
    \cline{2-5}
      ~ & Top-1 & Top-5 & Top-1 & Top-5\\
     \hline
     All & 41.1 & 65.8 & 11.0 & 23.6\\
     1000 $m$ & 49.2 & 76.7 & 19.9 & 41.5 \\
     500 $m$ & 54.1 & 82.6 & 26.4 & 53.3\\
     200 $m$ & 60.9 & 90.6 & 37.7 & 72.0\\
    \hline
    
    \hline
    \end{tabular}
    \caption{Retrieval accuracy (percentage) of the proposed method with noisy GPS signals.}
    \label{tab:noisy}
\end{table}
\vspace{-0.4cm}
\section{Conclusion}
We propose a new benchmark for cross-view image geo-localization beyond one-to-one retrieval, which is a more realistic setting for real-world applications. Our end-to-end framework first coarsely localizes the query with retrieval, and then refines the localization by predicting the offset with regression. An IOU-based hybrid loss is designed to leverage the supervision of semi-positive samples. Extensive results show great potential of the proposed method in realistic settings. Our proposed dataset offers a new testbed for cross-view geo-localization and inspires novel research in this field. 

{\textbf{Acknowledgement}. This work is partially supported by the National Science Foundation under Grant No. 1910844.}

{\small
\bibliographystyle{ieee_fullname}
\bibliography{egbib}
}

\end{document}